\documentclass[conference]{IEEEtran}
\IEEEoverridecommandlockouts
\usepackage{cite}
\usepackage{amsmath,amssymb,amsfonts}
\usepackage{textcomp}
\usepackage[utf8]{inputenc} % allow utf-8 input
\usepackage[T1]{fontenc}    % use 8-bit T1 fonts
\usepackage[bookmarks=false]{hyperref}
\usepackage{url}            % simple URL typesetting
\usepackage{booktabs}       % professional-quality tables
\usepackage{nicefrac}       % compact symbols for 1/2, etc.
\usepackage{microtype}      % microtypography
\usepackage[x11names]{xcolor}         % colors
\usepackage{adjustbox}
\usepackage{algorithm}
\usepackage{array}
\usepackage{algpseudocode}
\usepackage{amsthm}
\usepackage[font=small,skip=5pt]{caption}
\usepackage{enumitem}
\usepackage{graphicx}
\usepackage{multirow}  
\usepackage{multicol}
\usepackage{mathtools}
\usepackage{svg}
\usepackage{subcaption}
\usepackage{pgfplots,lipsum}
\usepgfplotslibrary{groupplots}
\usepackage{tikz,tikzscale}
\usetikzlibrary{patterns}
\usepackage{wrapfig}

\algnewcommand\algorithmicinput{\textbf{Input:}}
\algnewcommand\Input{\item[\algorithmicinput]}
\algnewcommand\algorithmicoutput{\textbf{Output:}}
\algnewcommand\Output{\item[\algorithmicoutput]}
\newcommand{\Best}[1]{\textcolor{Green4}{#1}}
\newcommand{\SBest}[1]{\textcolor{DarkOrange3}{#1}}

\def\BibTeX{{\rm B\kern-.05em{\sc i\kern-.025em b}\kern-.08em
    T\kern-.1667em\lower.7ex\hbox{E}\kern-.125emX}}
\begin{document}

\title{FedCC: Towards Addressing Label Distribution Skews in Distillation-Based Federated Learning\\
}
% https://anonymous.4open.science/r/FedCC-29EC/Readme.md

\author{
\IEEEauthorblockN{Wenxuan Ye \IEEEauthorrefmark{1}\IEEEauthorrefmark{2},
Onur Ayan\IEEEauthorrefmark{1}, 
Xueli An\IEEEauthorrefmark{1}, 
Georg Carle\IEEEauthorrefmark{2}}
\IEEEauthorblockA{
\textit{\IEEEauthorrefmark{1} Huawei Heisenberg Research Center, Huawei Technologies Duesseldorf GmbH}\\
\textit{\IEEEauthorrefmark{2} TUM School of Computation, Information and Technology, Technical University of Munich}
}
wenxuan.ye@tum.de, \{onur.ayan, xueli.an\}@huawei.com, carle@net.in.tum.de
}

\maketitle

\begin{abstract}
Federated Learning (FL) enables distributed clients to collaboratively train models without sharing raw data, making it promising for leveraging massive devices in communication networks.
In distillation-based FL, each client applies its local model on an \textit{unlabeled public dataset}, and shares only prediction results with the server.
While heterogeneous local data introduces \textit{label distribution skew}, thus biasing client models toward majority classes and leading to potentially inaccurate predictions.
The lack of ground‑truth labels in the public dataset hampers the server’s ability to calibrate predictions, which ultimately degrades overall performance.
To address this, we propose \textit{FedCC}, a simple and effective algorithm for mitigating client misclassification.
Instead of being forced to classify and risking error propagation, clients are allowed to \textit{tag ambiguous samples as `unknown'}. 
This additional class, together with calibrated pseudo‑labels on the public data, balances confidence in majority classes against uncertainty in under-represented ones.
Extensive experiments demonstrate that FedCC significantly outperforms existing methods, especially under severe label skew.
In the extreme scenario where each client holds samples from only one of ten classes, FedCC achieves 67.3\% accuracy, while baselines collapse to near‑random results.
% 
% 贡献需要说的更为宏观一点？
% Idea很好，所以一定要把价值写出来！
\end{abstract}


\section{Introduction}
\label{sec:intro}
% 参考“Federated Learning with Label-Masking Distillation” 写introduction。
Federated Learning (FL) \cite{mcmahan2017communication} enables collaborative model training across decentralized clients without exposing their raw data.
In the classical parameter‑based approach, each client refines the global model with its private data, uploads the updated parameters to the server, and the server aggregates these updates to produce the next‑round global model.
While effective, this approach incurs heavy communication overhead on parameter transmission, and assumes a homogeneous model architecture, overlooking the diverse bandwidth, computation, and data distributions across real‑world devices \cite{kairouz2021advances, ye2023advancing, liu2026foundation, bi2025llava}.

Drawing on Knowledge Distillation (KD) \cite{hinton2015distilling}, Distillation-based FL offers an alternative for client collaboration \cite{ye2026select}.
Each client runs its local model on a public dataset and uploads only the resulting prediction vectors (i.e., class‑score logits); 
the server then aggregates these vectors into a global prediction vector and redistributes it, reducing communication overhead by orders of magnitude.
In addition, because logits are architecture‑agnostic, the approach natively supports heterogeneous models, enabling flexible deployment across varied networked devices.
Regarding the public dataset, various strategies have been employed, including the use of a labeled dataset \cite{li2022not, Bi2025PRISMSI}, or the generation of synthetic data derived from local datasets or local model parameters \cite{dai2024enhancing}.
Recent studies \cite{lin2020ensemble} have relaxed the requirement by adopting unlabeled ones, avoiding costly labeling processes and alleviating privacy concerns.
Accordingly, this paper adopts \textit{an unlabeled public dataset} as an integral component of the scenario setting.

\begin{figure}
    % Center the figure.
    \begin{center}
    \includegraphics[width=0.85\columnwidth]{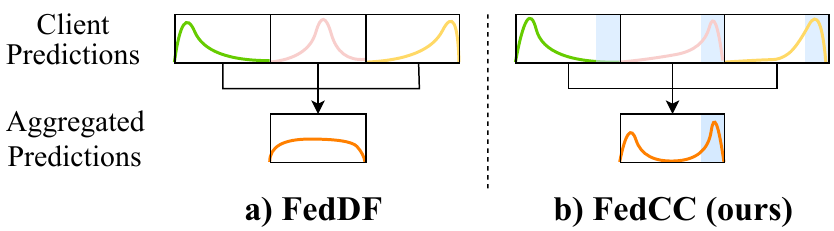}
    % \includesvg[width=\columnwidth]{fig/illu.svg}
    \captionsetup{font=small}
    \caption{
    Comparison of distillation-based FL under label distribution skews. 
    a) FedDF \cite{lin2020ensemble} averages biased local predictions, leading to suboptimal outcomes. 
    b) Our FedCC introduces an `unknown' class (highlighted in blue shadow), allowing clients to acknowledge their unknownness and thereby yielding more informative aggregation.
    }
    \label{fig:illu}
    \end{center}
\end{figure}

Despite its promise, distillation‑based FL is vulnerable to \textit{label distribution skew}: 
heterogeneous client data cause local models to overfit the majority classes, and the resulting biases are reflected in the prediction vectors \cite{kairouz2021advances}.
The challenge is further compounded by the absence of ground-truth labels in the public dataset, which prevents the server from calibrating logits and results in distorted supervision that degrades global performance.
Existing approaches in parameter-based FL (e.g., weighted model aggregation \cite{wang2020tackling}, model regularization \cite{li2020federated}) presume access to client model weights. 
This assumption breaks down in distillation‑based FL, where only logit vectors are exchanged, providing far sparser signals.
Recent efforts focus on weighting client updates by estimated reliability during server‐side aggregation \cite{gong2021ensemble}, moving beyond naive averaging as in FedDF\cite{lin2020ensemble}.
Others aim at amplifying minority-class information and aligning local models with the global objective \cite{fan2024federated}. 
Nevertheless, label-skew bias still creeps in, and existing schemes cannot consistently filter out misleading updates during aggregation.
Empirical evidence \cite{diao2023towards} demonstrates that, under label-distribution skew, local models overwhelmingly predict their majority classes, thereby injecting errors into the aggregated prediction (as in Fig.~\ref{fig:illu}a).

In response, we propose \textit{FedCC}, a simple and effective algorithm designed to mitigate \textit{C}lient mis\textit{C}lassification.
Treating the mismatch between a client’s skewed data and the ideal balanced distribution as an open‑set problem \cite{scheirer2012toward}, we introduce an `unknown' class to absorb minority and missing classes.
Instead of being forced to issue a class prediction and risking error propagation, clients are allowed to acknowledge their limitations and mark ambiguous inputs as `unknown'.
Illustrated in Fig.~\ref{fig:illu}b, this `unknown' class (shown in blue shadow) enables clients to express uncertainty, yielding more informative aggregated predictions.
To implement this, we incorporate the unlabeled dataset into the local objective function and employ a calibrated pseudo-label generation method, to weigh confidence in majority classes with the uncertainty hidden in unknown classes.
This weight is adaptively adjusted per client and per sample, optimizing the model’s generalization under diverse data conditions.

To evaluate performance, we conduct experiments on three widely used datasets: CIFAR-10, CIFAR-100, and TinyImageNet, across various label skew scenarios.
FedCC consistently outperforms state-of-the-art FL methods by at least 3\% points in most scenarios, and the accuracy gap grows as label skew becomes more severe. 
In the extreme scenario where each client holds samples from only one of ten classes, FedCC achieves an accuracy of 67.3\%, while other baselines fail to generate meaningful predictions.
Beyond delivering strong global generalization, FedCC enhances local model performance on its minority classes by leveraging uncertainty-aware predictions. 

To summarize, our key contributions are as follows:
\begin{itemize}[left=1em]
\item {
We introduce FedCC, a simple and effective method for mitigating client misclassification. 
We frame label skew as an open-set recognition problem, and extend the existing class framework with an `unknown' class to absorb minority and missing classes.
}
\item {
FedCC incorporates the unlabeled dataset into the local objective function, and proposes a novel pseudo-label generation method, enabling the model to recognize and account for unknownness.
This method can be interpreted as adding a regularizer, mitigating overfitting.
}
\item {
Extensive experiments demonstrate that FedCC significantly outperforms current FL methods, and the performance gap increases as label skew becomes more severe.
}
\end{itemize}

\section{Related Works}

\label{sec:related}

\noindent \textbf{Parameter-based FL with label distribution skews:}
While FL safeguards privacy by keeping data local, the skewed label distributions across clients undermine model generalization.
To stabilize heterogeneous training, methods like FedProx \cite{li2020federated} and FedSAM \cite{qu2022generalized} employ regularization, variance reduction, or gradient filtering.
Other approaches target class imbalance, e.g., via rescaling logits (FedRS \cite{li2021fedrs}), or calibrating predictions (FedLC \cite{zhang2022federated}, FedMR \cite{fan2024federated}, FedVLS \cite{guo2025exploring}).
Alternatively, FedConcat \cite{diao2024exploiting} departs from model averaging by concatenating local encoders to collaboratively train a global classifier.
While FedOV \cite{diao2023towards} uses open-set recognition to alleviate label skew, its reliance on synthesizing outlier samples with implicit labels limits its applicability to unlabeled data.

\noindent \textbf{Distillation-based FL with label distribution skews: }
To curb label skew under distillation‑based FL, FedNed \cite{lu2023federated} emphasizes distilling minority‐class knowledge, LightTS \cite{campos2023lightts} re-weights clients by strength, and Co-Boosting \cite{dai2024enhancing} alternates synthetic data generation with ensemble updates.
However, their reliance on labeled public datasets or large-scale synthesis raises annotation and privacy concerns.
When public data is unlabeled, the lack of ground-truth labels complicates bias correction. Existing solutions address this by retaining high-confidence predictions \cite{li2022not}, clustering clients \cite{deng2023hierarchical}, or communicating only hard labels \cite{abourayya2025little}. 
While emphasizing reliable local knowledge, these methods struggle to filter misinformation stemming from inherent local biases (As shown in Fig.~\ref{fig:illu}a). Thus, effective solutions for fair client-side classification remain an open challenge.

\section{Scenario Setting}
Consider an FL scenario with $K$ clients and $C$ classes (label set $\mathbb{C} = \{1, \dots, C\}$). 
Each client $k$ holds a private local dataset $\mathbb{D}_{l}^{(k)} = \{(x, y)\}$. 
Additionally, a global unlabeled public dataset $\mathbb{D}_u = \{x_u\}$ is accessible to all clients and the server.
Due to label skew, client label priors $P^{(k)}(y)$ vary, though we assume the conditional distribution $P(y|x)$ is consistent across clients \cite{kairouz2021advances}. 
We partition $\mathbb{C}$ into three mutually exclusive subsets ($\mathbb{C}_{J}^{(k)} \cup \mathbb{C}_{N}^{(k)} \cup \mathbb{C}_{S}^{(k)} = \mathbb{C}$) for each client $k$: majority classes $\mathbb{C}_{J}^{(k)}$ (sufficiently represented), minority classes $\mathbb{C}_{N}^{(k)}$ (sparse, noisy estimates), and missing classes $\mathbb{C}_{S}^{(k)}$ (absent).

\noindent \textbf{Collaborative distillation protocol:}
Client $k$ deploys a heterogeneous classifier $f^{(k)}(\cdot;\theta^{(k)})$. 
In each communication round, client $k$ computes logits $f^{(k)}(x_u) \in \mathbb{R}^{C}$ for public samples $x_u \in \mathbb{D}_u$. 
These are converted to probabilities $\hat{\mathbf{y}}^{(k)}(x_u) := \sigma^{(k)}(x_u)$ via a temperature-scaled softmax (with temperature $\tau=1$ omitted for brevity).

The server aggregates these probabilities into an ensemble teacher:
\begin{equation}
\hat{\mathbf{y}}^{(s)}(x_u) := \sum\nolimits_k \mu_k \, \hat{\mathbf{y}}^{(k)}(x_u), \quad \mu_k \geq 0, \; \sum\nolimits_k \mu_k = 1.
\end{equation}

Clients then synchronize their models by minimizing the Kullback-Leibler divergence with the teacher distribution:
\begin{equation}
\label{eq:ed}
\mathbb{E}_{x_u \sim \mathbb{D}_u} \left[\text{D}_{\text{KL}}\left( \hat{\mathbf{y}}^{(s)}(x_u) \parallel \sigma^{(k)}(x_u) \right)\right].
\end{equation}
Post-synchronization, clients refine their models on local private data before sharing predictions in the next round.

\section{Approach}
\label{sec:approach}
FedCC targets client‑side misclassification, and by lowering local errors, it ultimately improves global generalization.
Sections \ref{sec:local}-\ref{sec:weight} detail the proposed client‑level algorithm, and Section~\ref{sec:pipeline} outlines the overall federated training pipeline.
In Sections \ref{sec:local}–\ref{sec:weight} we focus on a single client, and therefore omit the client index for notational brevity (e.g., $f$, $\theta$ and $\sigma$ instead of $f^{(k)}$, $\theta^{(k)}$ and $\sigma^{(k)}$).

\subsection{Local learning objective}
\label{sec:local}
\noindent{\textbf{Semi-supervised learning:}}
We begin by reviewing Semi-Supervised Learning (SSL), a paradigm that leverages both labeled and unlabeled data to build effective models.
%, compared to supervised learning which solely relies on the labeled data.
%
A classical SSL strategy is self-training \cite{yarowsky1995unsupervised}, which bootstraps a model with its own high‑confidence predictions.
Firstly, a classifier $f(\cdot\, ;\theta_0)$ is fitted on labeled data $\mathbb{D}_{l} = \{(x, y)\}$.
The trained model $\theta_0$ then assigns pseudo-labels to the unlabeled data $x_u \in \mathbb{D}_{u}$ by selecting the class of maximum posterior probability: 
$\hat{y}(x_u) = \arg \max_{c \in \mathbb{C}} f_c(x_u;\theta_0)$.
Merging the original and pseudo-labeled examples yields an enlarged dataset on which the model is re‑optimized. 
The overall training objective is to minimize the following loss function: 
\begin{equation}
\label{eq:ssl}
\begin{aligned}
\mathcal{L}(\theta) := & \: \mathbb{E}_{(x, y) \sim \mathbb{D}_{l}} \left[\ell_{\text{CE}}(f(x;\theta), y)\right] \\
& +\lambda \: \mathbb{E}_{x_{u} \sim \mathbb{D}_{u}} \left[\, \ell_{\text{CE}}(f(x_u;\theta), \hat{y}(x_u)) \, \right]
\end{aligned}
\end{equation}
where the cross-entropy loss $\ell_{\text{CE}}(f(x), y) = -\log \sigma_y(x)$ quantifies the discrepancy between the label $y$ and prediction logits $f(x)$, and $\sigma_y(x)$ is given as the probability after the softmax operation; $\lambda > 0$ balances the labeled and unlabeled terms.

\noindent{\textbf{Our method: }}
As described in the scenario setting, client $k$ maintains a labeled dataset $\mathbb{D}_{l}^{(k)}$ and unlabeled dataset $\mathbb{D}_{u}$, mirroring the typical SSL structure.
To optimally utilize both datasets, we design the objective function based on Eq.~\ref{eq:ssl} as: 
\begin{equation}
\label{eq:objective}
\begin{aligned}
\mathcal{L}(\theta) := & \: \underbrace{\mathbb{E}_{(x, y) \sim \mathbb{D}_{l}^{(k)}} \left[\ell_{\text{CE}}(f (x;\theta), y)\right]}_{: \mathcal{L}_{l}(\theta)} \\
& +\lambda  \: \underbrace{\mathbb{E}_{x_u \sim \mathbb{D}_{u}} \left[\ell_{\text{CE}}(f(x_u;\theta), \hat{y}(x_u))\right]}_{: \mathcal{L}_{u}(\theta)}
\end{aligned}
\end{equation}
where $\mathcal{L}_{l}(\theta)$ and $\mathcal{L}_{u}(\theta)$ denote the losses on the labeled and unlabeled datasets, respectively.

As introduced before, client $k$ first fits a model $\theta_{0}$ on its labeled data, detailed as
$\theta_{0} := \arg \min\nolimits_{\theta} \mathcal{L}_{l}(\theta)$.
Ideally, $\theta_{0}$ would master classes it has encountered, i.e., $\mathbb{C}_{J}^{(k)} \bigcup \mathbb{C}_{N}^{(k)}$.
However, in practice, the severe under-representation of minority classes $\mathbb{C}_{N}^{(k)}$ leads to poorly calibrated posterior probabilities, which often drift toward majority classes \cite{zhang2022federated}. 
Furthermore, $\theta_{0}$ lacks the ability to recognize samples from missing classes as it has never learned them during training.

\textbf{To mitigate misclassification of minority and missing classes, we enlarge the label space with an auxiliary `unknown' class $u$, which acts as a catch-all for samples with unreliable predictions.}
With the label space augmented to $\mathbb{C} \, \cup \, \{u\}$, the model then satisfies
$\sum_{y \in \mathbb{C}} \sigma_{y}(x_u) + \sigma_{u}(x_u) = 1$.

Building on this, we ideally partition the unlabeled dataset $\mathbb{D}_u$ into samples from majority classes $\mathbb{C}_{J}^{(k)}$, and all others.
If the sample’s true class belongs to client $k$'s majority set $y(x_u) \in \mathbb{C}_{J}^{(k)}$, the pseudo-label is assigned to: $m(x_u): = \arg\max_{c \in \mathbb{C}_{J}^{(k)}}(f_c(x_u;\theta_{0}))$, as abundant local examples enable the client to identify these classes reliably \cite{zou2018unsupervised};
if $y(x_u) \not\in \mathbb{C}_{J}^{(k)}$, the sample is intended to be classified as $u$.
Then, the ideal pseudo-label assignment is
\begin{equation}
    \hat{y}(x_u) = {m(x_u)}  \, \text{ if } \,  y(x_u) \in \mathbb{C}_{J}^{(k)} \, \text{ else } \, u
\end{equation}
The unlabeled loss $\mathcal{L}_{u}(\theta)$ is therefore reformulated as:  
\begin{equation}
\begin{aligned}
\mathbb{E}_{x_u \sim \mathbb{D}_u} \big[ &\, \mathbf{1}[y(x_u) \in \mathbb{C}_{J}^{(k)}] \: \ell_{\text{CE}}(f(x_u;\theta), m(x_u)) \\ 
& + (1 - \mathbf{1}[y(x_u) \in \mathbb{C}_{J}^{(k)}]) \: \ell_{\text{CE}}(f(x_u;\theta), u) \, \big]
\end{aligned}
\end{equation}
where the indicator $\mathbf{1}[P]$ equals 1 if the event $P$ is true and 0 otherwise.
Since the truth labels $y(x_u)$ are unavailable, we approximate this indicator with a soft weight $\alpha(x_u) \in [0, 1]$:
\begin{equation}
\label{eq:lu}
\begin{aligned}
\mathcal{L}_{u}(\theta) \approx \mathbb{E}_{x_u \sim \mathbb{D}_u} \big[ &\, \alpha(x_u) \, \ell_{\text{CE}}(f(x_u;\theta), m(x_u)) \\ 
& + (1 - \alpha(x_u)) \, \ell_{\text{CE}}(f(x_u;\theta), u) \, \big]
\end{aligned}
\end{equation}
The comparison between Eq.~\ref{eq:objective} and Eq.~\ref{eq:lu} yields the pseudo-label in probabilistic-vector form:
\begin{equation}
\label{eq:pseudo}
\hat{\mathbf{y}}(x_u) := \alpha(x_u) \cdot \mathbf{e}_{m(x_u)} + (1-\alpha(x_u)) \cdot \mathbf{e}_{u}
\end{equation}
where $\mathbf{e}_{m(x_u)}$ is the one-hot vector for the predicted class $m(x_u)$, and $\mathbf{e}_{u}$ represents that of class $u$.
This pseudo-label distributes probability mass over only two classes: the predicted class $m(x_u)$ and the unknown class $u$.
We detail the concrete design of $\alpha$ in the following subsection.

\subsection{Weight design}
\label{sec:weight}

% After training on the local dataset $\mathbb{D}_{l}^{(k)}$, client obtains the preliminary model $\theta_{0}^{(k)}$.
% Our method accommodates probabilities for learned classes and integrates the likelihood of unlearned classes.
% $\alpha(x_u)$ accounts for class prior discrepancies arising from label distribution skew, balancing confidence in learned classes against uncertainty from unlearned ones. 
The weight $\alpha(x_u)$ re-weighs the local loss on unlabeled data so that learning is guided by global rather than local class prevalence.
Instead of relying on class probabilities, which risk privacy leakage, we derive the weights solely from entropies, since they are inexpensive to compute and reflect uncertainty.

\begin{algorithm}[htbp]
\caption{Algorithm of FedCC pipeline}
\label{alg:fedcc}
\begin{algorithmic}[1]
\Require $K$ clients, global rounds $T$, local labeled dataset $\{\mathbb{D}_l^{(k)}\}_{k=1}^K$, public unlabeled dataset $\mathbb{D}_u$.
\Statex \textbf{Training:} 
\For{$t = 1$ to $T$}
  \Statex \textbf{Client $k$ (in parallel):}
  \State Obtain model $\theta_{sync}^{(k),t}$ by solving the optimization problem specified in Eq.~\ref{eq:ed}. (Omit when $t=1$)
  \State $\theta_{0}^{(k), t} \gets \arg\min\nolimits_{\theta^{(k)}}\mathcal{L}_{l}(\theta^{k})$  
 \Comment{Initialize at $\theta_{sync}^{(k),t}$}
  \State \textcolor{DarkOrchid3}{$\theta^{(k), t} \gets \textsc{FedCC\_Local}(\theta_{0}^{(k), t}, \mathbb{D}_{l}^{(k)}, \mathbb{D}_{u})$}
  \State Compute pseudo-predictions:  
  \Statex \quad $\hat{\mathbf{y}}^{(k),t}(x_u) \gets \sigma_{\tau}^{(k)}(x_u;\theta^{(k),t})\;\forall\,x_u\in\mathbb{D}_u$
  \State Upload $\{\hat{\mathbf{y}}^{(k),t}(x_u)\}$ to server
  
  \Statex \textbf{Server:}
  \State $\mu^{(k),t}(x_u) \gets \text{normalize } \{1 - \hat{y}_u^{(k), t}(x_u)\}_k$
  \State  $\tilde{\mathbf{y}}^{(s), t}(x_u) \gets \sum\nolimits_{k} \mu^{(k),t}(x_u) \, \hat{\mathbf{y}}^{(k), t}(x_u)$
  \State \textcolor{DarkOrchid3}{Set the unknown-class entry of $\tilde{\mathbf{y}}^{(s),t}(x_u)$ to zero, and renormalize the remaining entries to get  
    $\hat{\mathbf{y}}^{(s), t}(x_u) \in \mathbb{R}^{C}$}
  \State Broadcast $\{\hat{\mathbf{y}}^{(s), t}(x_u)\}$ back to all clients
\EndFor
\State \Return $\{\theta^{(k),T}\}_{k=1}^K$

\Procedure{FedCC\_Local}{$\theta_{0}^{(k),t},\mathbb{D}_l^{(k)},\mathbb{D}_u$}
  \State Generate pseudo-labels via Eq.~\ref{eq:pseudo} with $\alpha^{(k),t}(x_u)$ from Eq.~\ref{eq:alpha}
  \State Obtain model $\theta^{(k), t}$ by optimizing Eq.~\ref{eq:objective}
  \State \Return $\theta^{(k), t}$
\EndProcedure

  \Statex
\Statex \textbf{Inference: Given a new input $x$} 
\State Generate $\tilde{\mathbf{y}}^{(s), T}(x_u)$ via Eq.~\ref{eq:mu_k}
\State $\hat{y} = \arg \max\nolimits_{c \in \mathbb{C}} \tilde{\mathbf{y}}^{(s), T}_{c}(x_u)$
\end{algorithmic}
\end{algorithm}

\noindent{\textbf{Weight towards majority classes ($w_l$):}}
Following \cite{lee2024cdmad}, we regard a solid-color image $x_0$ as featureless, and probe the classifier’s intrinsic bias by feeding it into the pre-trained model, recording
$\mathbf{p}^{(k)} := \sigma(x_{0};\theta_{0})$.
To keep our entropy‑based weighting scheme responsive even for near‑certain predictions, we apply label smoothing \cite{yuan2020revisiting}:
\begin{equation}
\label{eq:p^k}
 \tilde{\mathbf p}^{(k)}=(1-\delta)\,\mathbf p^{(k)}+\delta\frac{\mathbf 1}{|\mathbb C|}   
\end{equation}
The weight is then
$w_l := H\bigl(\tilde{\mathbf p}^{(k)}\bigr)$, 
where $H(\cdot)$ denotes entropy.
The smoothed entropy quantifies how strongly the model already favors its local label set and stays strictly positive even under a one-class bias.

\noindent{\textbf{Weight over unknown classes ($w_u$):}} 
Classes under-represented by client $k$ are collected in $\mathbb{C}_{N}^{(k)} \bigcup \mathbb{C}_{S}^{(k)}$ with number being $|\mathbb{C}| - |\mathbb{C}_{J}^{(k)}|$.
Since the client lacks prior knowledge for those classes, we assign them the maximum uncertainty, defined as the entropy of a uniform distribution. 
This yields
$w_{u} := \log(|\mathbb{C}| -|\mathbb{C}_{J}^{(k)}| + 1)$.
Here, we add one virtual category to account for residual ambiguity introduced by the majority classes, reflecting that the client remains uncertain even with only one under-represented class.

Since $w_l$ quantifies the client’s concentration of probability over majority classes and $w_u$ captures its uncertainty about unknown classes, a natural way to trade off these two effects is through their ratio:
$\alpha_{base} = \frac{w_l}{w_l + w_u} \in [0, 1]$.
Note that $w_l$ and $w_u$ are client-specific and vary across clients.

To further account for \textbf{sample-specific prediction uncertainty}, we quantify the pretrained model’s confidence for each unlabeled instance $x_u$ by $w_{x_u}:= \min\big(w_l, H( \sigma(x_{u};\theta_{0}))\big)$.
As high entropy indicates low confidence in assigning the sample to a class in $\mathbb{C}$, we penalize $\alpha_{base}$ with $w_{x_u}$, leading to the final $\alpha(x_u)$ expression:
\begin{equation}
\label{eq:alpha}
\alpha(x_u) :=  \frac{w_l - w_{x_u}}{w_l + w_u}
\end{equation}

\textbf{This weighting factor $\alpha$ limits the extent to which a local model’s bias affects pseudo‑labeling.}
$\alpha$ reserves a non‑zero probability for the `unknown' class, allowing the model to consider an alternative label rather than forcing a majority‑class assignment. 
By deferring these uncertain cases, $\alpha$ curbs early mislabelling of minority samples and prevents such errors from cascading through subsequent training rounds.
% The coefficient $\alpha(x_u)$ quantifies the model's confidence in assigning the sample to a majority class, while captures the uncertainty contained in minority and missing classes. 

After generating pseudo-labels, each client updates its local model via Eq.~\ref{eq:objective} to obtain $\theta$, which is then used to generate predictions shared with the server.

\subsection{Federated training pipeline}
\label{sec:pipeline}
The training process runs over $T$ communication rounds.
For notational clarity, we use the superscript $(k)$ and $(s)$ to denote the client index and the server-aggregated information, respectively, and $t$ to indicate the communication round.

At the beginning of round $t$, each client $k$ downloads the aggregated soft targets from the previous round, denoted as 
$\hat{y}^{(s),t-1}(x_u), \forall x_u \in \mathbb{D}_u$, and then use Eq.~\ref{eq:ed} to synchronize its local model.
Then each client adapts its model to its private labeled dataset $\mathbb{D}_{l}^{(k)}$, resulting in the intermediate model $\theta_{0}^{(k), t}$.
Unlike prior distillation-based FL schemes that share predictions immediately, FedCC inserts an extra refinement step.
Each client generates the pseudo-label via Eq.~\ref{eq:pseudo} and obtains the final round-specific parameters $\theta^{(k), t}$ by optimizing the objective in Eq.~\ref{eq:objective}.
Each client then evaluates all samples in $\mathbb{D}_u$ and produces temperature-scaled prediction vectors, $\hat{\mathbf{y}}^{(k), t}(x_u) = \sigma_{\tau}^{(k)}(x_u;\theta^{(k), t}) \in \mathbb{R}^{C+1}$, where the last entry corresponds to the unknown class.
These prediction vectors are uploaded to the server.
Clients' contributions are weighted by known-class confidence:
we take $1 - \hat{y}_u^{(k), t}(x_u)$ as a preliminary weight, then normalize it across clients to produce $\mu^{(k), t}(x_u)$.
The server then aggregates the received predictions as 
\begin{equation}
\label{eq:mu_k}
\tilde{\mathbf{y}}^{(s), t}(x_u) = \sum\nolimits_k \mu^{(k), t}(x_u) \: \hat{\mathbf{y}}^{(k), t}(x_u)
\end{equation}
It then zeroes out the unknown-class component and renormalizes the remaining $C$ entries to yield a valid probability vector $\hat{\mathbf{y}}^{(s), t}(x_u) \in \mathbb{R}^{C}$.
The resulting soft targets are then redistributed to all clients to initiate round $t+1$.

During inference, we adopt the ensemble strategy used in FedMD \cite{li2019fedmd} and FedOV \cite{diao2023towards}.
After completing all $T$ rounds, clients transmit their model $f^{(k)}$ with parameters $\theta^{(k), T}$ to the server.
The server then aggregates the predictions from all client models using Eq.~\ref{eq:mu_k} and selecting the most probable label among the original classes $\mathbb{C}$. 
Formally, the predicted label is given by $
\hat{y} = \arg \max\nolimits_{c \in \mathbb{C}} \tilde{\mathbf{y}}^{(s), T}_{c}(x_u)$.

Algorithm~\ref{alg:fedcc} details the complete FedCC procedure.

\section{Experiments}
\label{sec:experiments}

\subsection{Experimental details}
\label{sec:details}

\renewcommand{\arraystretch}{0.4}
\begin{table*}[htbp]
\centering
\small
\setlength{\tabcolsep}{3pt}
\vspace*{0.05in}
\caption{Performance comparison under various label skews. 
FedCC demonstrates superior performance across almost all scenarios, and an increasing performance gap as label skew intensifies.}
\label{tab:multi}
% \hspace*{-2.5cm}
\begin{adjustbox}{width=0.95\textwidth}
\begin{tabular}{c|cc|cc|cc|cc|cc|cc}
\toprule
\multirow{3}{*}{\textbf{Method}} & \multicolumn{4}{c|}{\textbf{TinyImageNet}} & \multicolumn{4}{c|}{\textbf{CIFAR-100}} & \multicolumn{4}{c}{\textbf{CIFAR-10}} \\
\cline{2-13}
 & \multicolumn{2}{c|}{$p \sim \text{Dir}(\eta)$} & \multicolumn{2}{c|}{$|\mathbb{C}_{J}^{(k)}|=N$} & \multicolumn{2}{c|}{$p \sim \text{Dir}(\eta)$} & \multicolumn{2}{c|}{$|\mathbb{C}_{J}^{(k)}|=N$} & \multicolumn{2}{c|}{$p \sim \text{Dir}(\eta)$} & \multicolumn{2}{c}{$|\mathbb{C}_{J}^{(k)}|=N$} \\
& $0.5$ & $0.05$ & $40$ & $20$ & $0.5$ & $0.05$ & $20$ & $10$ & $0.5$ & $0.05$ & $2$ & $1$ \\
\midrule
FedAvg & $36.2_{\pm1.6}$ & $24.1_{\pm1.3}$ & \SBest{$31.1_{\pm1.9}$} & $14.3_{\pm0.8}$ & $\SBest{49.3_{\pm1.2}}$ & $34.2_{\pm2.1}$ &  $33.5_{\pm1.3}$ & $18.2_{\pm1.3}$ & $72.6_{\pm2.6}$ & $44.3_{\pm4.2}$ & $37.5_{\pm1.0}$ & $11.0_{\pm1.1}$ \\
FedProx & $35.0_{\pm3.0}$ & $26.4_{\pm1.0}$ & $28.4_{\pm0.5}$ & $13.1_{\pm2.0}$  & $48.9_{\pm0.9}$ & $29.3_{\pm0.7}$ & $30.6_{\pm1.8}$ & $16.1_{\pm0.6}$ & $70.1_{\pm2.3}$ & $47.2_{\pm2.0}$ & $41.0_{\pm1.3}$ & $10.6_{\pm0.4}$ \\
FedNova & $34.9_{\pm1.1}$ & $25.7_{\pm0.6}$ & $27.2_{\pm1.3}$ & $14.1_{\pm1.0}$ & $47.8_{\pm1.7}$ & $33.7_{\pm1.7}$& $37.9_{\pm1.5}$ & $20.9_{\pm1.1}$ & $75.3_{\pm1.1}$ & $42.5_{\pm0.7}$ & $35.2_{\pm0.8}$ & $10.0_{\pm0.0}$ \\
FedRS & $32.5_{\pm1.3}$ & $22.1_{\pm1.0}$ & $27.7_{\pm1.4}$ & $12.9_{\pm1.5}$ & $46.3_{\pm0.8}$ & $27.8_{\pm0.9}$ & $34.2_{\pm1.4}$ & $17.7_{\pm1.2}$ & $\Best{76.6_{\pm1.9}}$ & $53.0_{\pm1.0}$ & $35.4_{\pm1.5}$ & $10.0_{\pm0.0}$\\
FedSAM & $35.5_{\pm0.8}$ & \SBest{$30.3_{\pm1.1}$} & $29.4_{\pm1.2}$ & $14.1_{\pm0.9}$ & $48.7_{\pm1.3}$ & $33.7_{\pm1.3}$ & $35.3_{\pm1.7}$ & $16.5_{\pm0.8}$ & $74.6_{\pm0.6}$ & $43.8_{\pm1.6}$ & $32.8_{\pm1.3}$ & $10.0_{\pm0.0}$ \\
FedLC & \SBest{$39.2_{\pm0.9}$} & $28.0_{\pm1.0}$ & $28.8_{\pm1.0}$ & $12.5_{\pm1.3}$ & $48.5_{\pm1.1}$ & \SBest{$34.6_{\pm0.7}$} & \SBest{$39.4_{\pm1.1}$} & $21.4_{\pm0.4}$ & $76.0_{\pm1.2}$ & $41.5_{\pm0.6}$ & $36.2_{\pm1.6}$ & $10.0_{\pm0.0}$ \\
FedMR & $34.8_{\pm0.9}$ & $24.7_{\pm0.5}$ & $29.2_{\pm0.8}$ & $15.1_{\pm1.3}$  & $\Best{50.1_{\pm1.2}}$ & $\SBest{34.6_{\pm0.9}}$ & $38.6_{\pm1.5}$ & $21.8_{\pm1.0}$ & \SBest{$76.2_{\pm0.9}$} & \SBest{$57.8_{\pm1.0}$} & $38.4_{\pm0.9}$ & $11.7_{\pm0.2}$ \\
% FedGF &&&& &&&& &&&&  \\
% FedCPD &&&& &&&& &&&& \\
FedConcat & $28.3_{\pm2.5}$ & $18.7_{\pm3.9}$ & $22.3_{\pm2.3}$ & $9.9_{\pm1.2}$ & $40.3_{\pm2.4}$ & $21.2_{\pm1.7}$ & $20.6_{\pm0.4}$ & $14.7_{\pm1.7} $ & $61.3_{\pm1.0}$ & $34.0_{\pm2.1}$ &  $28.5_{\pm1.2}$& \SBest{$12.7_{\pm0.7}$} \\
FedVLS & $26.4_{\pm1.4}$ & $20.7_{\pm0.4}$ & $22.4_{\pm0.7}$ & $15.3_{\pm2.6}$ & $43.5_{\pm1.6}$ & $27.3_{\pm1.2}$ & $27.1_{\pm3.5}$ & $18.3_{\pm0.3}$ & $66.0_{\pm2.8}$ & $39.5_{\pm0.8}$ & $32.9_{\pm3.6}$ & $10.0_{\pm0.0}$ \\
\midrule
FedMD & $30.1_{\pm1.7}$ & $24.8_{\pm1.5}$ & $30.2_{\pm0.8}$ & \SBest{$23.1_{\pm1.6}$} & $36.4_{\pm0.7}$ & $28.1_{\pm0.7}$ & $34.9_{\pm1.2}$ & \SBest{$26.4_{\pm1.6}$} & $69.0_{\pm0.7}$ & $34.2_{\pm2.6}$ & \SBest{$54.5_{\pm1.0}$} & $10.7_{\pm0.9}$ \\
FedDF & $26.8_{\pm2.0}$ & $16.9_{\pm2.7}$ & $20.7_{\pm0.6}$ & $10.3_{\pm2.8}$ & $33.6_{\pm1.2}$ & $22.1_{\pm0.8}$ & $27.6_{\pm3.1}$ & $14.1_{\pm1.1}$ & $63.9_{\pm1.9}$ & $39.9_{\pm3.7}$ & $50.8_{\pm1.4}$ & $11.7_{\pm1.0}$ \\
LSR & $26.2_{\pm0.9}$ & $16.7_{\pm1.1}$ & $23.8_{\pm1.2}$ & $10.2_{\pm0.6}$ &  $31.8_{\pm1.6}$ & $21.3_{\pm0.8}$ & $27.8_{\pm1.4}$ & $16.5_{\pm0.3}$& $58.6_{\pm1.0}$ & $31.4_{\pm0.7}$ & $27.2_{\pm1.4}$& $10.0_{\pm0.0}$\\
FedET & $23.8_{\pm3.8}$ & $18.5_{\pm2.0}$ & $21.3_{\pm0.8}$ & $11.6_{\pm1.0}$ & $27.2_{\pm0.9}$ & $22.3_{\pm1.7}$ & $28.0_{\pm1.1}$ & $22.1_{\pm0.8}$ & $62.1_{\pm2.0}$ & $38.4_{\pm2.7}$ & $51.2_{\pm0.6}$ & $10.0_{\pm0.0}$ \\
FedHKT & $27.7_{\pm1.0}$ & $21.3_{\pm0.9}$ & $26.9_{\pm1.7}$ & $14.6_{\pm0.4}$ & $34.4_{\pm1.9}$ & $25.6_{\pm0.5}$ & $31.8_{\pm3.6}$ & $21.3_{\pm0.9}$ & $67.6_{\pm1.4}$ & $44.9_{\pm0.9}$ & $49.1_{\pm0.4}$ & $11.1_{\pm1.2}$ \\
FedCT & $25.9_{\pm0.5}$ & $17.7_{\pm1.3}$ & $21.2_{\pm0.7}$ & $9.9_{\pm2.6}$ & $33.2_{\pm0.4}$ & $21.8_{\pm1.6}$ & $26.7_{\pm0.8}$ & $12.2_{\pm2.3}$ & $65.0_{\pm1.2}$ & $29.7_{\pm1.7}$ & $24.1_{\pm0.2}$ & $10.0_{\pm0.0}$\\
\midrule
FedCC & $\Best{40.6_{\pm0.8}}$ & $\Best{34.8_{\pm1.0}}$ & $\Best{36.9_{\pm2.0}}$ & $\Best{34.3_{\pm1.3}}$ 
& $48.9_{\pm1.3}$ & $\Best{40.5_{\pm1.8}}$ & $\Best{45.3_{\pm1.4}}$ & $\Best{42.6_{\pm1.3}}$ 
& $74.7_{\pm1.6}$ & $\Best{60.9_{\pm2.2}}$ & $\Best{68.2_{\pm1.1}}$ & $\Best{67.3_{\pm1.8}}$ \\
\bottomrule
\end{tabular}
\end{adjustbox}
\end{table*}
\renewcommand{\arraystretch}{0.4}
\begin{table}[t!]
\centering
\small
\setlength{\tabcolsep}{3pt}
\caption{Local model performance over local and global test data, where FedCC leads in almost every case. (CIFAR-100 subset)}
\label{tab:local}
\begin{adjustbox}{width=0.95\linewidth} % 缩减了宽度比例以适应更少的列
\begin{tabular}{c|c|cc|cc}
\toprule
\textbf{Test data} & \textbf{Local model} & \multicolumn{2}{c|}{$p \sim \text{Dir}(\eta)$} & \multicolumn{2}{c}{$|\mathbb{C}_{J}^{(k)}| = N$} \\
& & $0.5$ & $0.05$ & $20$ & $10$ \\
\midrule
& FedAvg & $\SBest{37.9_{\pm2.2}}$ & $60.0_{\pm2.7}$ & $\SBest{57.4_{\pm3.1}}$ & $\SBest{72.3_{\pm4.9}}$ \\
\multirow{4}{*}{Local} & FedLC & $36.1_{\pm1.4}$& $\SBest{60.5_{\pm4.4}}$ & $56.9_{\pm5.1}$ & $72.2_{\pm6.0}$ \\
& FedMR & $36.7_{\pm2.2}$ & $60.2_{\pm5.1}$ & $56.7_{\pm4.7}$ & $70.6_{\pm5.3}$ \\
& SSL & $34.6_{\pm2.0}$ & $58.0_{\pm2.9}$ & $54.6_{\pm2.0}$ & $70.5_{\pm5.2}$ \\
& FedCC & $\Best{38.2_{\pm2.3}}$ & $\Best{64.4_{\pm2.6}}$ & $\Best{60.0_{\pm4.6}}$ & $\Best{76.6_{\pm3.3}}$ \\
\midrule
& FedAvg & $17.6_{\pm1.4}$ & $10.1_{\pm0.9}$ & $11.3_{\pm0.8}$ & $7.2_{\pm0.6}$ \\
\multirow{4}{*}{Global} & FedLC & $19.0_{\pm1.0}$ & $\SBest{11.0_{\pm1.6}}$ & $\SBest{11.8_{\pm1.0}}$ & $7.0_{\pm0.8}$ \\
& FedMR & $\SBest{19.3_{\pm1.5}}$ & $10.7_{\pm1.7}$ & $11.6_{\pm1.1}$ & $7.0_{\pm0.8}$ \\
& SSL & $15.4_{\pm1.1}$ & $10.2_{\pm1.3}$ & $11.4_{\pm0.9}$ & $\SBest{7.3_{\pm0.5}}$ \\
& FedCC & $\Best{71.1_{\pm0.3}}$ & $\Best{74.7_{\pm1.2}}$ & $\Best{71.1_{\pm2.4}}$ & $\Best{77.9_{\pm3.8}}$ \\
\bottomrule
\end{tabular}
\end{adjustbox}
\end{table}

\noindent\textbf{Dataset settings:}
Following the previous work \cite{zhang2022federated, guo2025exploring}, we conduct classification tasks using CIFAR-10, CIFAR-100 and TinyImageNet datasets.
We randomly divide each dataset into two parts: a public dataset and a private dataset.
The public dataset, which comprises $10\%$ of the total training samples, has all labels removed and is shared among all participants. 
To simulate different label skews in the private dataset, we use two partition methods: 
(1) Dirichlet distribution $p \sim \text{Dir}(\eta)$: for each class $c$, we sample $p_c \sim \text{Dir}_{K}(\eta)$, then assign client $k$ a fraction $p_{c,k}$ of the class-$c$ samples to its private set. 
A lower $\eta$ value indicates greater heterogeneity.
(2) Pathological partition $|\mathbb{C}_{J}^{(k)}| = N$: each client has data from a fixed number of classes $N$, with an equal number of samples for each class distributed among the assigned clients.
% equal distribution of samples among assigned clients. 

\noindent\textbf{Baseline:}
We benchmark FedCC against nine parameter-based methods (FedAvg \cite{mcmahan2017communication}, FedProx \cite{li2020federated}, FedNova \cite{wang2020tackling}, FedRS \cite{li2021fedrs}, FedSAM \cite{qu2022generalized}, FedLC \cite{zhang2022federated}, FedMR \cite{fan2024federated}, FedConcat \cite{diao2024exploiting}, FedVLS \cite{guo2025exploring}) and six distillation-based methods (FedMD \cite{li2019fedmd}, FedDF \cite{lin2020ensemble}, LSR \cite{yuan2020revisiting}, FedET \cite{cho2022heterogeneous}, FedHKT \cite{deng2023hierarchical}, FedCT \cite{abourayya2025little}). 
For fairness, we exclude methods requiring a labeled public dataset or those based on generating labeled data.
For details, please refer to the corresponding paper.

\noindent\textbf{Experiment settings:}
Inspired by \cite{deng2023hierarchical}, parameter-based FL uses ResNet20 on all nodes; in distillation-based FL, clients split evenly between ResNet14 and ResNet20.
The setup includes $10$ clients, $15$ global epochs, and $20$ local epochs per step in client model training.
We adopt Adam optimizer with batch size $64$, learning rate $0.001$.
Temperature $\tau$ is set to $1$, tradeoff $\lambda$ in Eq.~\ref{eq:objective} to $0.2$, $\delta$ in $\tilde{\mathbf p}^{(k)}$ (Eq.~\ref{eq:p^k}) to $10^{-3}$.
We categorize classes by client $k$'s bias $\mathbf{p}^{(k)}$ (Eq.~\ref{eq:p^k}): a class $c$ is a majority if $p^{(k)}_{c} > 2\%$, minority if $0 < p^{(k)}_{c} \leq 2\%$, and missing if $p^{(k)}_{c} =0$.
We report results over three random seeds, with model accuracy shown without percent signs.

\subsection{An overall comparison}
Experiment results, as summarized in Table~\ref{tab:multi}, show that FedCC outperforms current FL methods across almost all tested scenarios, with its advantage widening as label skew intensifies.
On heavy skew ($p \sim \text{Dir}(0.05)$ or pathological partition), it is ahead by at least 3\% points and often by more than 10\%.
In the harshest scenario where each client has access to only one class of CIFAR‑10, baselines collapse to near‑random guessing ($\leq$ 12.7\%), yet FedCC still delivers a robust 67.3\%.
This robustness stems from permitting clients to only predict familiar data and to label uncertain samples as `unknown', thereby abstaining from unreliable predictions and reducing misleading information.

% \noindent \textbf{Comparison of local model performance:} 
% We evaluate local model performance on the CIFAR-100 dataset against various baselines across both local and global test data (Table~\ref{tab:local}). On local data, FedCC leverages uncertainty to significantly boost minority-class accuracy without sacrificing majority-class performance. On global data, FedCC effectively prevents error propagation and improves overall performance by correctly classifying under-represented samples as the `unknown' class.

\noindent \textbf{Comparison of local model performance:}
We evaluate the performance of local models on both local and global test data, as shown in Table~\ref{tab:local}.
The models are generated using FedAvg \cite{mcmahan2017communication}, FedLC \cite{zhang2022federated}, FedMR \cite{fan2024federated}, standard SSL \cite{yarowsky1995unsupervised}, and our FedCC.
For the global test data, if the true label does not belong to the majority classes, classifying it as the `unknown' class is considered valid.
As expected, FedCC shows significant improvement on the global test data.
On the local test set, FedCC leverages uncertainty to supply richer signals for minority classes and increases their likelihood of being correctly learned, consistently outperforming all baselines.

\subsection{Performance visualization}

% \begin{figure}[t!]
% \centering
% \begin{subfigure}[b]{0.22\textwidth}
%         \centering
%         \includegraphics[width=\textwidth]{fig/Single Client 4 on CIFAR-10 Model.png} 
%         \caption{FedAvg local model}
%         \label{fig:single}
% \end{subfigure}%
% \hspace{0.25em}% Space between image A and B
% \begin{subfigure}[b]{0.22\textwidth}
%         \centering
%         \includegraphics[width=\textwidth]{fig/Model Avg.png}  
%         \caption{FedAvg global model}
%         \label{fig:avg}
% \end{subfigure}%
% \\
% % \hspace{0.25em}% Space between image A and B 
% % \vspace{0.5em}
% \begin{subfigure}[b]{0.22\textwidth}
%         \centering
%         \includegraphics[width=\textwidth]{fig/Extend Client 4 on CIFAR-10 Model.png} 
%         \caption{FedCC local model}
%         \label{fig:extended}
% \end{subfigure}%
% \hspace{0.25em}% Space between image B and I_c
% \begin{subfigure}[b]{0.22\textwidth}
%         \centering
%         \includegraphics[width=\textwidth]{fig/FedCP_15.png} 
%         \caption{FedCC aggregation}
%         \label{fig:FedCC}
% \end{subfigure}%
% \caption {
% T-SNE visualizations of learned representations on CIFAR-10 where the local distribution follows $|\mathbb{C}_{J}^{(k)}=1|$.
% % reference: MOON (Model-Contrastive Federated Learning)
% }
% \label{fig:cifar10}
% \end{figure}

\begin{figure}[t!]
\centering
\begin{subfigure}[b]{0.24\textwidth}
        \centering
        \includegraphics[width=\textwidth]{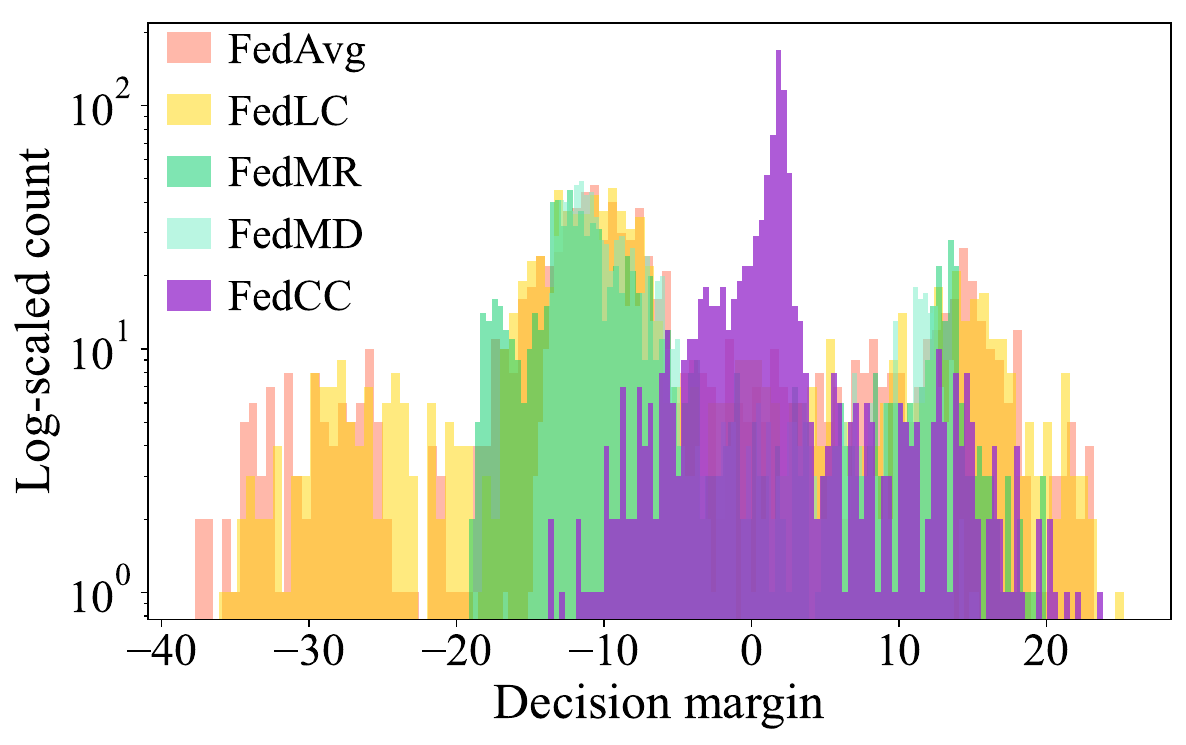} 
        \caption{$p \sim \text{Dir}(0.05)$}
        \label{fig:diri0.05}
\end{subfigure}%
\hspace{0.25em}% Space between image A and B
\begin{subfigure}[b]{0.23\textwidth}
        \centering
        \includegraphics[width=\textwidth]{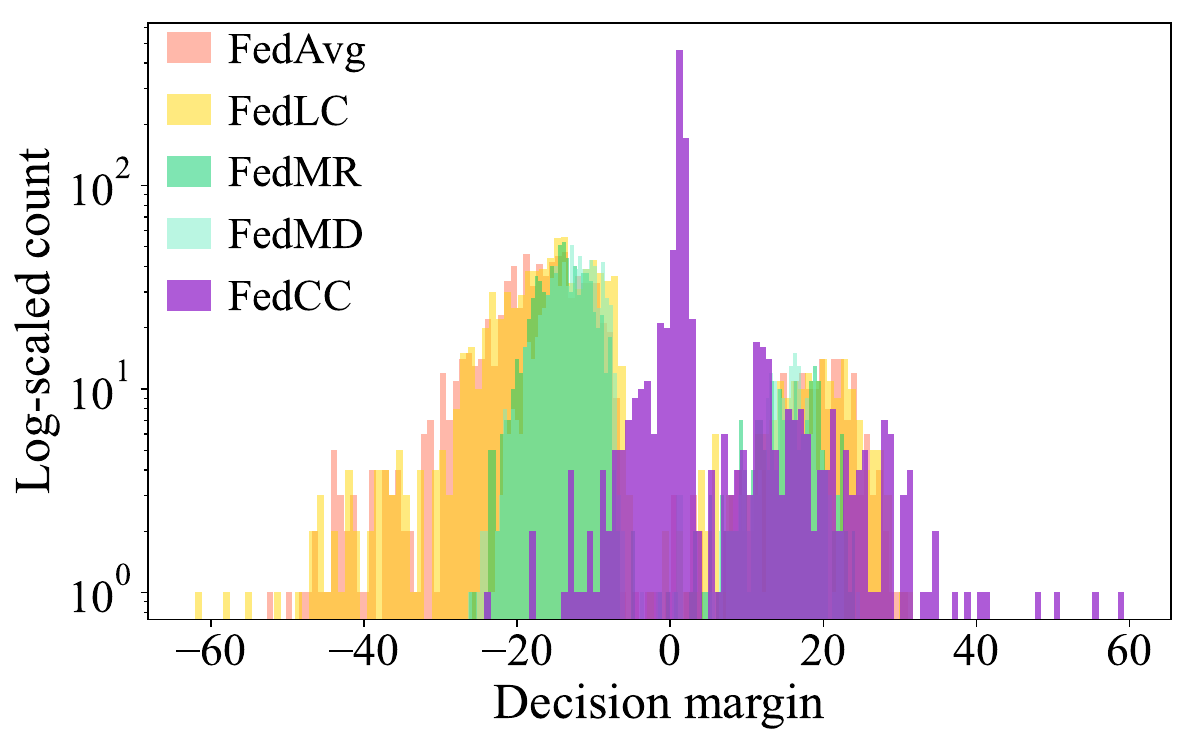}  
        \caption{$|\mathbb{C}_{J}^{(k)}|=2$}
        \label{fig:class2}
\end{subfigure}%
\caption {
Distribution of decision margins for local model on CIFAR‑10.
% reference: MOON (Model-Contrastive Federated Learning)
}
\label{fig:margin}
\end{figure}

\begin{figure}[t!]
\centering
\begin{subfigure}[b]{0.12\textwidth}
        \centering
        \includegraphics[width=\textwidth]{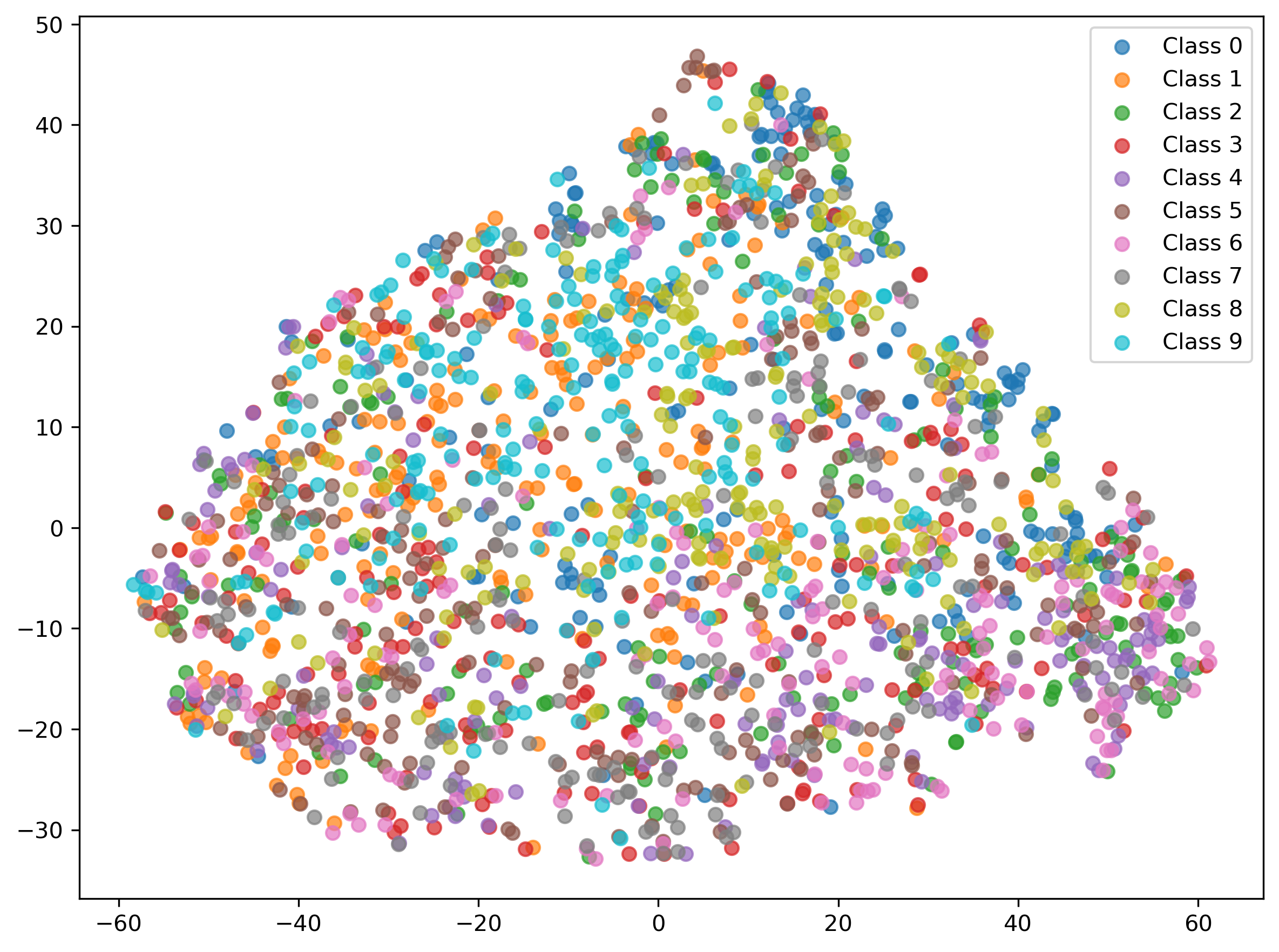} 
        \caption{FedAvg local}
        \label{fig:single}
\end{subfigure}%
% \hspace{0.1em}% Space between image A and B
\begin{subfigure}[b]{0.12\textwidth}
        \centering
        \includegraphics[width=\textwidth]{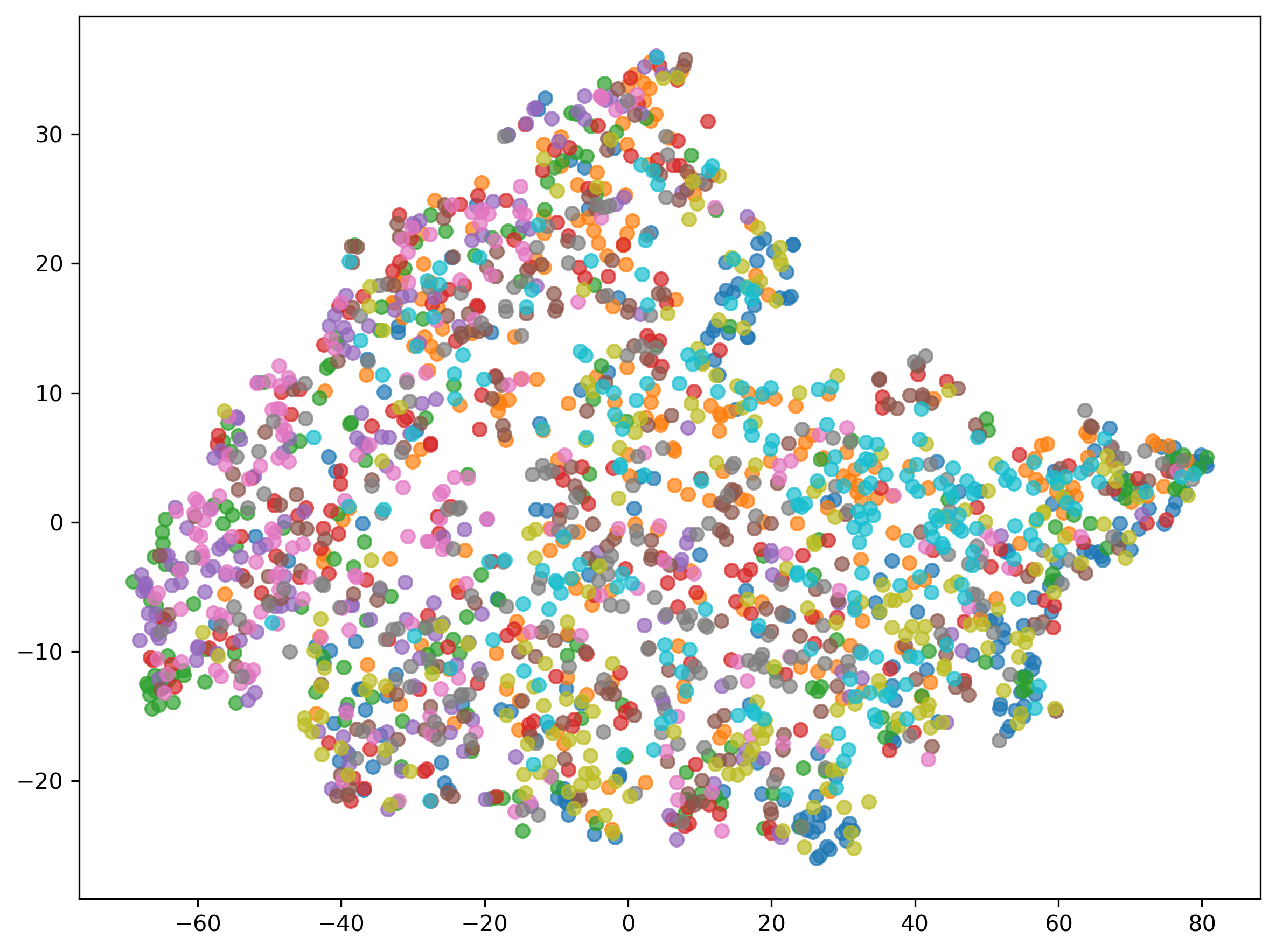}  
        \caption{FedAvg global}
        \label{fig:avg}
\end{subfigure}%
% \hspace{0.1em}% Space between image A and B 
% \vspace{0.5em}
\begin{subfigure}[b]{0.12\textwidth}
        \centering
        \includegraphics[width=\textwidth]{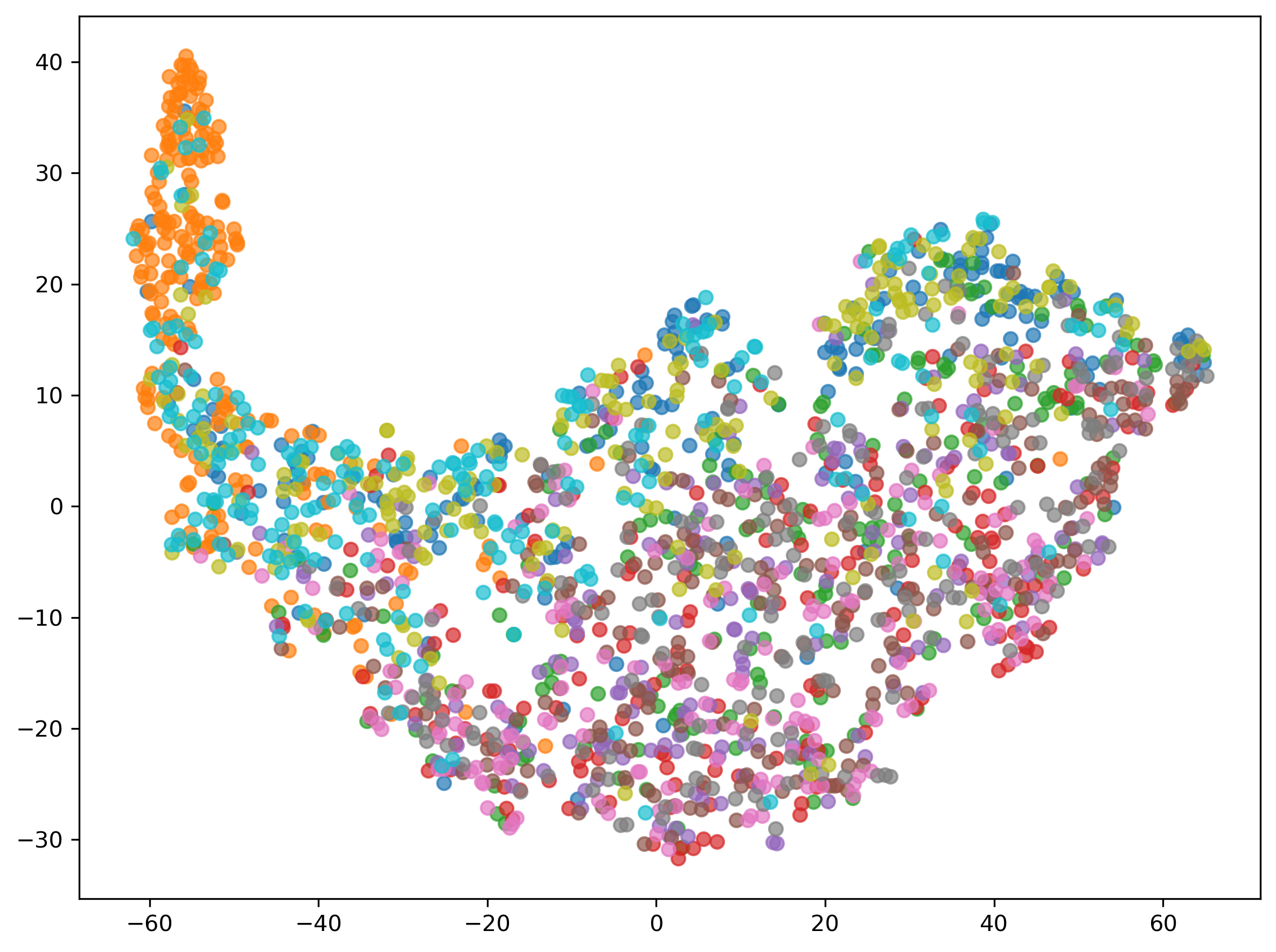} 
        \caption{FedCC local}
        \label{fig:extended}
\end{subfigure}%
% \hspace{0.1 em}% Space between image B and I_c
\begin{subfigure}[b]{0.12\textwidth}
        \centering
        \includegraphics[width=\textwidth]{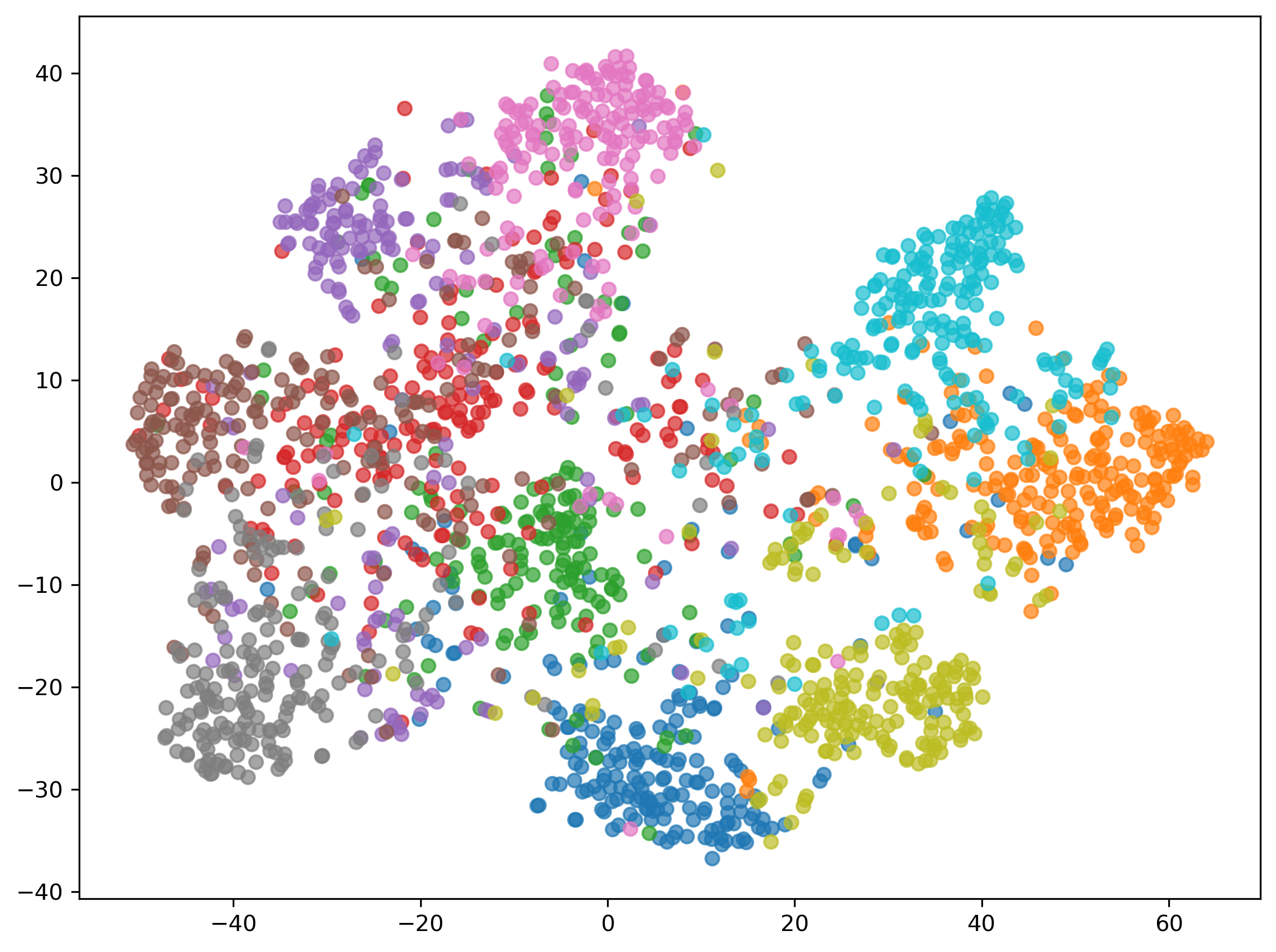} 
        \caption{FedCC global}
        \label{fig:FedCC}
\end{subfigure}%
\caption {
T-SNE visualizations of learned representations on CIFAR-10 where the local distribution follows $|\mathbb{C}_{J}^{(k)}=1|$.
% reference: MOON (Model-Contrastive Federated Learning)
}
\label{fig:cifar10}
\end{figure}

\noindent \textbf{Decision margin:}
The decision margin measures the gap between the model's score for the correct class and its highest incorrect guess.
Evaluating under two challenging data partitions, Figure~\ref{fig:margin} shows margin histograms with a logarithmic y-axis.
FedCC markedly shifts the distribution rightward to achieve a positive mean, whereas the best competing method remains negative.
This rightward shift confirms that FedCC effectively ensures clear class separation and suppresses client-side misclassification, even under severe label skew.

\noindent \textbf{T-SNE analysis of learned representations:}
Figure~\ref{fig:cifar10} visualizes the learned feature representations where each client holds data from a single class ($|\mathbb{C}_{J}^{(k)}| = 1$) on CIFAR‑10, where each color denotes a distinct class.
The embeddings are obtained by applying PCA whitening followed by t-SNE in a unified pipeline.
\textit{Baseline}: 
each client trains a local model using standard cross-entropy on its data, with the embedding of a representative client shown in Fig.\ref{fig:single}, and that of the FedAvg global model in Fig.\ref{fig:avg}.
Due to the lack of inter-class variation, local representations collapse along class‑specific axes, and averaging these misaligned feature spaces yields weak global performance.
\textit{FedCC}:
we next train local models with FedCC’s objective and aggregate them by ensembling their soft predictions.
The corresponding client‑side embedding (Fig.\ref{fig:extended}) cleanly isolates the client’s own class from the “unknown” region, and the aggregated embedding (Fig.\ref{fig:FedCC}) recovers distinct clusters for all classes.
FedCC thus curbs error propagation by explicitly separating learned versus unseen classes at the client level, allowing the server to combine information without blurring minority categories.

\section{Conclusion}
\label{sec:conclusion}
We present FedCC, a novel distillation-based FL algorithm that mitigates client misclassification by incorporating an auxiliary `unknown' class.
By enabling clients to defer ambiguous predictions to this class, FedCC acts as an implicit regularizer, preventing overconfidence on majority classes and suppressing noisy updates.
Extensive empirical evaluation confirms that it consistently outperforms state-of-the-art baselines in both local and global performance.
Crucially, FedCC’s lightweight reliance on logit exchanges and robust performance make it well suited for deployment in real‑world communication networks, which involves large numbers of devices with heterogeneous compute power, bandwidth, and data distributions.
% Ablation studies further validate the contributions of our design components.
% , and one‑shot communication experiments demonstrate its robustness under stringent computation and communication constraints

{
    \small
    \bibliographystyle{IEEEtran}
    \bibliography{IEEEabrv}
}
\end{document}